\documentclass[runningheads]{llncs}

\newcommand{\tinyshrinker}{\vspace{-0.2cm}}

\usepackage{graphicx}
\usepackage{enumitem}
\setenumerate[1]{itemsep=0pt,partopsep=0pt,parsep=\parskip,topsep=5pt}
\setitemize[1]{itemsep=0pt,partopsep=0pt,parsep=\parskip,topsep=5pt}
\setdescription{itemsep=0pt,partopsep=0pt,parsep=\parskip,topsep=5pt}
\usepackage{array}
\usepackage{multirow}
\usepackage{booktabs}
\usepackage[normalem]{ulem}
\usepackage{mathrsfs}
\graphicspath{{figs/}}
\usepackage{float} 
\usepackage{amsmath,amssymb,amsfonts,bm}
\useunder{\uline}{\ul}{}
\usepackage{pgfplots} 
\usepackage{tikz}
\usepackage{amsmath,bm}
\usepackage{algorithm}
\usepackage{algorithmic}
\usepackage[misc]{ifsym}
\usepackage{lipsum}
\pgfplotsset{width=9cm,compat=1.9}
\makeatletter
\def\blfootnote{\xdef\@thefnmark{}\@footnotetext}
\makeatother

\begin{document}
%
% \title{Contribution Title\thanks{Supported by organization x.}}
\title{A Behavior-aware Graph Convolution Network Model for Video Recommendation}

\titlerunning{A Behavior-aware GCN Model for Video Recommendation}
\author{Wei Zhuo\inst{1}
\and
Kunchi Liu\inst{2,3}
\and
Taofeng Xue\inst{2,3}
\and
Beihong Jin\inst{2,3(\textrm{\Letter})}
\and
Beibei Li\inst{2,3}
\and
 Xinzhou Dong\inst{2,3}
\and
He Chen\inst{1}
\and
Wenhai Pan\inst{1}
\and
Xuejian Zhang\inst{1}
\and
Shuo Zhou\inst{1}
}

% \author{}

\authorrunning{W. Zhuo et al.}

\institute{MX Media Co., Ltd., Singapore, Singapore \and State Key Laboratory of Computer Science, Institute of Software, Chinese Academy of Sciences, Beijing, China\\ \email{Beihong@iscas.ac.cn}\\
\and University of Chinese Academy of Sciences, Beijing, China}

\blfootnote{This work was supported by the National Natural Science Foundation of China under Grant No. 62072450 and the 2019 joint project with MX Media.}

% \let\thefootnote\relax\footnotetext{This work was supported by the National Natural Science Foundation of China under Grant No. 62072450 and the 2019 joint project with MX Media.}


\maketitle
% typeset the header of the contribution
%
\vspace{-0.5cm}
\begin{abstract}
Interactions between users and videos are the major data source of performing video recommendation. Despite lots of existing recommendation methods, user behaviors on videos, which imply the complex relations between users and videos, are still far from being fully explored. In the paper, we present a model named Sagittarius. Sagittarius adopts a graph convolutional neural network to capture the influence between users and videos. In particular, Sagittarius differentiates between different user behaviors by weighting and fuses the semantics of user behaviors into the embeddings of users and videos. Moreover, Sagittarius combines multiple optimization objectives to learn user and video embeddings and then achieves the video recommendation by the learned user and video embeddings. The experimental results on multiple datasets show that Sagittarius outperforms several state-of-the-art models in terms of recall, unique recall and NDCG. 

\end{abstract}
\vspace{-0.8cm}
\section{Introduction}
% \tinyshrinker
In recent years, online streaming platforms develop rapidly. On the mainstream online streaming platforms (e.g., YouTube and MX Player), the number of daily active users can easily reach hundreds of millions, where users are constantly changing their needs of cultural entertainment and tastes. In order to improve user experience and increase user stickiness to platforms, video recommendation becomes an indispensable part of the online streaming platforms. 

Video recommendation, in common with other item recommendation tasks, has to face a large amount of sparse user-video interactions which contain multiple user behaviors such as clicking, sharing and downloading videos. Further, we note that the behaviors in the video recommendation more or less indicate user preferences, although their roles differ from ones in the e-commerce scenarios where some behaviors such as place-an-order clearly reveal the user preferences. 

In general, the interactions between users and videos are the most important data source for top-k video recommendation, since they imply user preferences on videos and prevailing trends of videos. To maintain the characteristics of original user-video interactions to the greatest extent, we plan to model interactions as a  bipartite user-video graph and learn embeddings of users and videos from the graph first, and then achieve the video recommendations. Logically, the videos interacted with a user can be used to enrich the user's embedding, because these interactions might reflect the user preference on the videos. On the other hand, a group of users who have interacted with a video can also be regarded as the side information of the video to measure the collaborative similarity between videos, therefore they can also be used to enrich the embedding of the video. For digging out the influence between users and videos from a large-scale sparse  bipartite user-video graph, we turn to a convolutional neural network on the graph to learn embeddings of users and videos. As a result, we propose a model named Sagittarius. The contributions of our work are summarized as follows.

\begin{itemize}
    \item [$\bullet$] We highlight that the different user behaviors imply the different degrees of user preferences on videos. To fully understand user preferences on videos, we quantify the behaviors as weights on the edges while building the bipartite graph, and then design a graph convolution network (GCN) to propagate the embeddings of users and videos across the edges of the bipartite user-video graph, so as to mine the influence between users and videos.
    \item [$\bullet$] We highlight that top-k video recommendation can be optimized from different ranking metrics. Further, we adopt a combination of multiple optimization objectives (including one major and two minor objectives) to guide the embedding learning. In particular, we propose to add weights of user behaviors to optimization objectives.
    \item [$\bullet$] We conduct extensive offline experiments and online A/B tests. The offline experimental results on five datasets show that Sagittarius outperforms several state-of-the-art models in terms of recall, unique recall and NDCG. Moreover, we conduct online A/B test in MX Player. Sagittarius behaves better than two existing models in MX Player, which proves the effectiveness of Sagittarius in a real-world production environment.
\end{itemize}

The rest of the paper is organized as follows. Section 2 introduces the related work. Section 3 describes our Sagittarius model. Section 4 gives the experimental results and analyses. Finally, the paper is concluded in Section 5.

\section{Related Work}
\tinyshrinker

Our work is related to the research under two non-orthogonal topics: GNN (Graph Neural Network)-based collaborative filtering, video recommendation.

GNN-based collaborative filtering combines representation learning on graphs \cite{hamilton2017representation} and collaborative filtering. The basic procedure is to model user-item interactions as one or multiple graphs, design a graph neural network (e.g., a GCN) to learn the node embeddings, and then apply the learned embeddings of items and users to achieve recommendation tasks.

Obviously, item graphs can be constructed from user-item interactions where nodes denote the items that users interact with and directed edges indicate the relations between two item nodes. For example, Equuleus \cite{xue2020feedback} constructs a homomorphic video graph  and then develops a node attributed encoder network to generate video embeddings. SR-GNN \cite{wu2019session} builds the directed item graphs for interaction sequences and then develops a GNN to capture complex item transition and an attention mechanism to fuse user’s long-term and short-term interests. GC-SAN \cite{xu2019graph}, an improved version of SR-GNN, borrows the self-attention structure from Transformer and combines a multi-layer self-attention network with original SR-GNN. 

Besides item graphs subordinated to homomorphic graphs, heteromorphic graphs can  be built. Taking the work on bipartite graphs as an example, Berg et al. present a graph auto-encoder GCMC \cite{berg2018graph}, where the encoder contains a graph convolution layer that constructs user and item embeddings through message passing on the bipartite user-item graph and the bilinear decoder predicts the labeled links in the graph. Wang et al. propose NGCF \cite{wang2019neural} which models the high-order connectivity on the  bipartite user-item graph. By stacking multiple embedding propagation layers, NGCF can generate the embeddings of users and items on the user-item graph, which encodes the collaborative signal between user and item. Wei et al. propose MMGCN \cite{wei2019mmgcn} for micro-videos, which constructs a  bipartite user-item graph for each modality, yields modal-specific embeddings of users and videos by the message passing of graph neural networks, enriching the representation of each node with the topological structure and features of its neighbors, and then obtains final node embeddings by a combination layer. Moreover, MBGCN \cite{jin2020multi} builds a unified heterogeneous graph where an edge indicates the behavior between a user and an item or the relation between items. In particular, the complex behavior-aware message propagations are designed for the GCN in MBGCN. 

Recently, for exploring the intrinsic relations of interaction data and/or fusing various side information, some work develops different types of graphs, such as semi-homogenous graphs in Gemini \cite{xu2020gemini}, the directed multigraph and the short-cut graph in LESSR \cite{chen2020handling}, the item graph, category graph, and shop graph in M2GRL \cite{wang2020m2grl}, and attribute graphs in Murzim \cite{dong2019improving}.

In addition, He et al. propose LightGCN \cite{he2020lightgcn}, a simplified version of NGCF, which deletes feature transformation and nonlinear activation in NGCF so as to decrease the unnecessary complexity of the network architecture. The authors claim that LightGCN can achieve better recommendation performance than NGCF. Similarly, Chen et al. propose LR-GCCF \cite{chen2020revisiting} which is a linear residual graph convolutional neural network based on graph collaborative filtering. By removing nonlinear transformations in the network, it reduces multiple parameter matrices of different layers of the network into a single matrix, thereby effectively reducing the amount of learnable parameters. 

As for video recommendation, recent progress mainly depends on deep learning. For example, Gao et al. \cite{gao2017unified} adopt recurrent neural networks and consider video semantic embedding, user interest modeling and user relevance mining in a unified framework. Li et al. \cite{li2019routing} present a model which is composed of a novel temporal graph-based LSTM and multi-level interest layers to model diverse and dynamic user interest and multi-level user interest. Moreover, Li et al. \cite{li2019long} employ GCNs to recommend long-tail hashtags for micro-videos.

Video recommendation is also a hot spot in the industry. For example, Baluja et al. \cite{baluja2008video} propose a random walk-based model for video click-through rate prediction in YouTube, Xu et al. \cite{xu2018hulu} adopt GATs (Graph Attention Networks) and the knowledge graph to generate video recommendations for the store-shelf and autoplay scenarios in Hulu, and Xue et al. \cite{xue2019spatio} develop a spatio-temporal collaborative filtering approach for offline on-demand cinemas of iQIYI.

Comparing to existing work, our Sagittarius model falls in the scope of GNN-based collaborate filtering, and designs a new graph convolution network to distinguish user behaviors and employs multiple optimization objectives. Further, differing from the recommendation models which exploits multi-behavior data, our model adopts a lightweight way to fuse user behavior semantics.

\section{Sagittarius Model}
\subsection{Problem Formulation}

In the top-k video recommendation scenario, we model user-video interactions as a  bipartite user-video graph, denoted by $\mathcal{G}=(\mathcal{U} \cup \mathcal{V},\mathcal{E},\mathcal{R})$, where $\mathcal{U}$ is the set of users, $\mathcal{V}$ is the set of videos, and $\mathcal{U} \cup \mathcal{V}$ constitutes the set of nodes. Edge $(u,v,r)$ in $\mathcal{E}$ represents that the type of the interaction between user $u \in \mathcal{U}$ and video $v \in \mathcal{V}$ is $r \in \mathcal{R}$, where $\mathcal{R}$ is a set of all interactive behavior types, including clicking, giving a like, sharing, downloading and etc. In addition, we define a priori function $\phi(r):\mathcal{R} \rightarrow \mathbb{R}$, which maps an interactive behavior to a score. The function is used to measure the degree of user preference by the user behavior, and higher scores indicate greater user interest. Particularly, the score of non-interaction is set to 0. While building the bipartite graph, we take the value of $\phi(r)$ as the weight $W_{uv}$ of edge $(u,v,r)$.
Our goal is to learn the embeddings of the users and videos and apply the embeddings to recommending top-k videos for users.

\subsection{Model Architecture}
\begin{figure}[H]
\centering 
\includegraphics[width=0.7\textwidth]{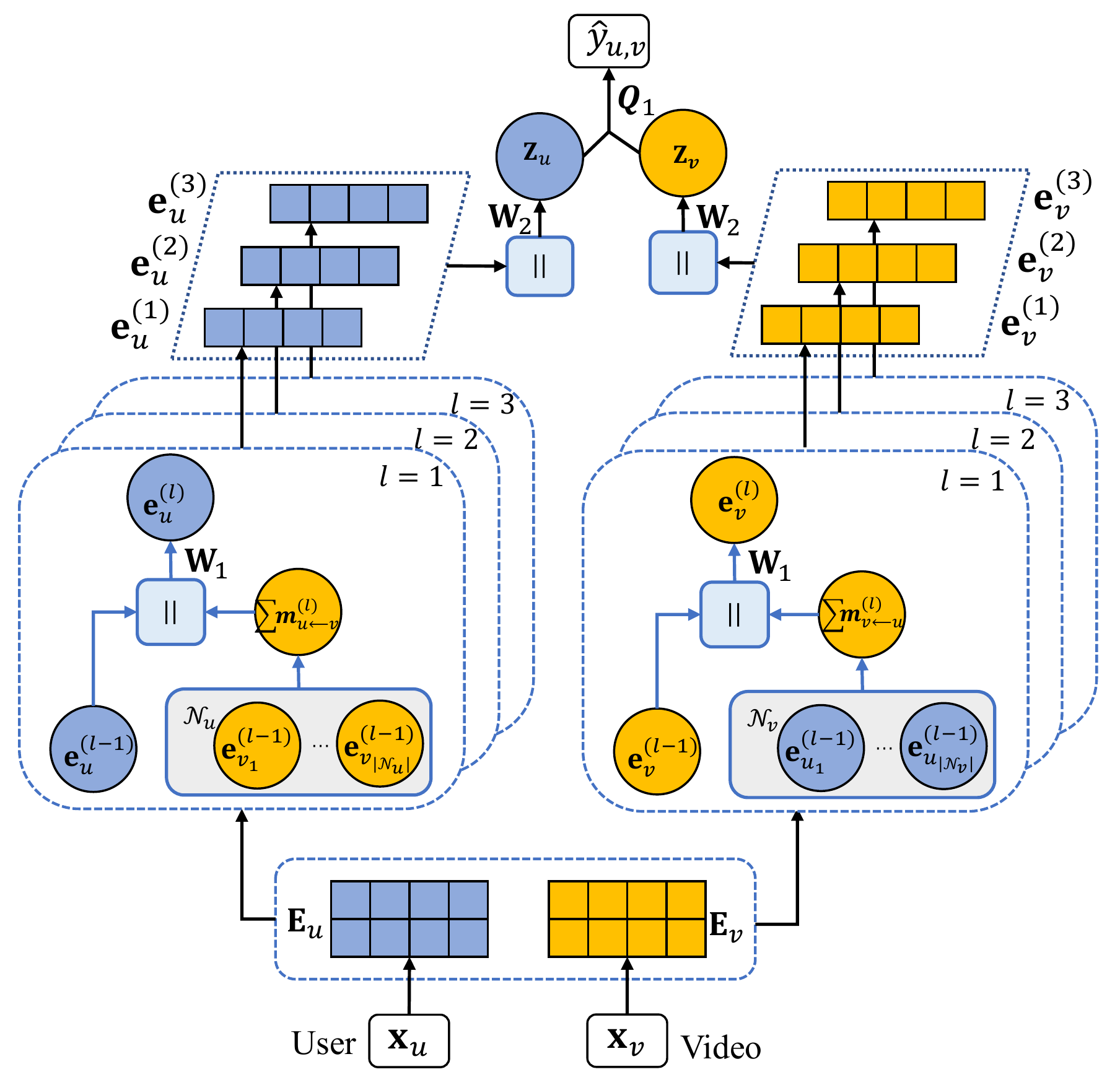}
\caption{The architecture of Sagittarius} 
\label{fig1}
\end{figure}

Sagittarius consists of the embedding layer, the convolution layers, the combination layer, and the prediction layer. Figure \ref{fig1} shows the architecture of our Sagittarius model.

\noindent\textbf{Embedding Layer} The embedding layer provides the initial embeddings of users and videos. Formally, we denote the user embedding matrix by $\mathbf{E}_u \in \mathbb{R}^{|\mathcal{U}|\times \bar{d}}$ and the video embedding matrix by $\mathbf{E}_v \in \mathbb{R}^{|\mathcal{V}|\times \bar{d}}$, where $\bar{d}$ denotes the initial embedding size. $\mathbf{E}_u$ and $\mathbf{E}_v$ are initialized randomly. In this way, given the user $u$ and its ID one-hot vector $\mathbf{x}_u \in \mathbb{R}^{|\mathcal{U}|}$, its embedding $\mathbf{e}_u \in \mathbb{R}^{\bar{d}}$ is set to $\mathbf{e}_u = \mathbf{E}_u^T  \mathbf{x}_u$. Similarly, for the ID one-hot vector $\mathbf{x}_v \in \mathbb{R}^{|\mathcal{V}|}$ of video $v$, its embedding is expressed as $\mathbf{e}_v = \mathbf{E}_v^T  \mathbf{x}_v$.

\noindent\textbf{Convolution Layers} Each convolution layer is responsible for performing the convolution operation on each node and its neighbors of the bipartite graph so that the embeddings will be passed and transformed across edges of the graph. In particular, for a certain user-video interaction pair $(u,v)$, in the $l$-th graph convolution layer, we propagate embedding information from the neighbor node $v$ to the target node $u$ by Equation \eqref{eq:muil}.

\begin{equation}
\mathbf{m}_{u \leftarrow v}^{(l)}= c_{uv}\mathbf{e}_v^{(l-1)}
\label{eq:muil}
\end{equation}

In Equation \eqref{eq:muil}, $\mathbf{m}_{u \leftarrow v}^{(l)}$ denotes the message passed from node $v$ to $u$, and $\mathbf{e}_v^{(l-1)}$ is the embedding of node $v$ output by the $(l-1)$-th convolution layer. In particular, $\mathbf{e}_v^{(0)}=\mathbf{e}_v$, $\mathbf{e}_v$ is the initial embedding of the video $v$ output by the embedding layer. $c_{uv}$ is the scaling factor, which is defined in Equation \eqref{eq:cui} where $\mathcal{N}_u$ and $\mathcal{N}_v$ are the sets of neighbors of node $u$ and node $v$, respectively. 
\begin{equation}
c_{uv}=\sqrt{\frac{\phi(r)}{\left|\mathcal{N}_u\right|\left|\mathcal{N}_v\right|}}\frac{{\mathbf{e}_{u}^{(l-1)}}^{T}\mathbf{e}_{v}^{(l-1)}}{||\mathbf{e}_{u}^{(l-1)}||_2||\mathbf{e}_{v}^{(l-1)}||_2}
\label{eq:cui}
\end{equation}

Further, we combine the aggregated neighbor information with the information of the node $u$ by Equations \eqref{eq:hul}-\eqref{eq:eul}.
\begin{equation}
\mathbf{h}_{u}^{(l)}= \mathbf{e}_u^{(l-1)} || \sum_{v\in \mathcal{N}_u} \mathbf{m}_{u \leftarrow v}^{(l)}
\label{eq:hul}
\end{equation}
\begin{equation}
\mathbf{e}_u^{(l)}={\rm ReLU}(\mathbf{W}_1\mathbf{h}_u^{(l)})
\label{eq:eul}
\end{equation}

In the above equations, $||$ denotes the concatenation operation, $\mathbf{e}_u^{(l-1)}$ is the embedding of the node $u$ output by the $(l-1)$-th convolution layer, and matrix $\mathbf{W}_1 \in \mathbb{R}^{\bar{d} \times 2\bar{d}}$ maps $\mathbf{h}_{u}^{(l)}$ to the $\bar{d}$-dimensional embedding space, which is a learnable parameter. By these equations, $\mathbf{e}_u^{(l-1)}$ is concatenated with the aggregated information $\sum_{v \in \mathcal{N}_u} \mathbf{m}_{u \leftarrow v}^{(l)}$, and then the concatenated result is sent to a nonlinear transformation, obtaining the user embedding of the current layer.

\noindent\textbf{Combination Layer} After the iterative operations of multiple convolution layers, we can obtain the output of each layer of the $L$ convolution layers, namely $\{\mathbf{e}_u^{(1)},\mathbf{e}_u^{(2)},\dots,\mathbf{e}_u^{(L)}\}$. In order to improve the expressiveness of the embeddings, we perform a layer-wise combination operation as follows.

\begin{equation}
\mathbf{h}_{u}= \mathbf{e}_u^{(1)} || \mathbf{e}_u^{(2)} || \cdots || \mathbf{e}_u^{(L)}
\label{eq:hu}
\end{equation}
\begin{equation}
\mathbf{z}_u=\mathbf{W}_2\mathbf{h}_u
\label{eq:zu}
\end{equation}

Specifically, we apply the concatenation operation shown in Equation \eqref{eq:hu} to concatenate the embeddings output by different layers, and perform the linear transformation shown in Equation \eqref{eq:zu} to combine the low-order embeddings extracted from the low-level convolution layers and the high-order embeddings extracted from the high-level convolution layers. The achieved effect is that the embeddings output by different layers can predict the score cooperatively. In Equation \eqref{eq:zu}, $\mathbf{W}_2 \in \mathbb{R}^{d \times L\bar{d}}$ maps $\mathbf{h}_{u}$ to a $d$-dimensional embedding space, where $d$ is the dimension of the embeddings. Thus, the finally learned $\mathbf{z}_u$ combines the information of the user node $u$ itself and the collaboration information from the multi-order neighborhood nodes.
In the same way, we can obtain the embedding  $\mathbf{z}_v$ of the video $v$.

\noindent\textbf{Prediction Layer}
We adopt the following bilinear decoder to calculate the affinity score of each user-video pair, sort videos by the score, and then filter the videos that the user has interacted with to get top-k recommendations.

\begin{equation}
\hat{y}_{u,v}=g_1(\mathbf{z}_u,\mathbf{z}_v)=\mathbf{z}_u^T \mathbf{Q}_1 \mathbf{z}_v
\label{eq:yui}
\end{equation}

In Equation \eqref{eq:yui}, $\mathbf{Q}_1$ is a parameter that needs to be learned, $\mathbf{z}_u$ and $\mathbf{z}_v$ are the embeddings of the node $u$ and the node $v$ obtained from Equation \eqref{eq:zu}.

\subsection{Learning Objectives}
Generally speaking, the recommended videos should be the top-k videos ranked by a certain metric or score. We choose the probability of the next click on each video as the metric, where the probability can be estimated based on the set of videos that user has interacted with or based on the sequence of videos that the user has interacted with. While employing a score, we can say that given a specific user, the score of the video that the user has interacted with should be greater than the score of the video that has not been interacted with. Therefore, we construct three objective functions, i.e., video partial order preserving function, video click-through rate prediction function and next click video prediction function.

\noindent\textbf{Video partial order preserving function}
In practical, we measure the partial order between videos instead of total order of videos, not only because the total order is often unavailable or costs high, but also because the partial order relationship between videos is enough to determine the relative ranking of recommended videos and alleviate the problem of positional deviation.

We adopt the BPR \cite{rendle2012bpr} loss function to optimize the partial order relationship between videos. Thus, for each user-video interaction $(u,v,r)$, i.e., one positive sample, we randomly select 10 negative samples to construct a set $\mathcal{O}$ containing 10 quadruple $(u,v,r,w)$, where user $u$ performs no behavior on video $w$. 

In addition, we regard the preference score difference between the positive sample and the negative sample, i.e. $W_{uv}-W_{uw}=\phi(r)-0=\phi(r)$ as the weight of this quadruple, where $\phi(r)$ is the function defined in Section 3.1.

Finally, we define the BPR loss function $\mathcal{L}_1$ for optimizing the ordering of videos as follows:

\begin{equation}
\mathcal{L}_1=-\frac{1}{\left|\mathcal{O}\right |} \sum_{\left (u,v,r,w \right )\in\mathcal{O}}\phi\left (r\right )\ln{\sigma\left(g_1\left(\mathbf{z}_u,\mathbf{z}_{v}\right)-g_1\left(\mathbf{z}_u,\mathbf{z}_{w}\right)\right)}
\label{eq:l1}
\end{equation}

In Equation \eqref{eq:l1}, $g_1\left(\mathbf{z}_u,\mathbf{z}_{v}\right)=\mathbf{z}_u^T \mathbf{Q}_1 \mathbf{z}_{v}$, and $\sigma$ is the sigmoid activation function. This function is our major learning objective. 

\noindent\textbf{Video click-through rate prediction function}
In the video recommendation scenario, most of the user-video interactions are very sparse and mainly focus on click behaviors. Considering that clicking can reflect the user's degree of interest in the video to a large extent, we recognize the necessity for optimizing the prediction of the click-through rate of users on the video. We adopt the binary cross entropy loss function to predict the video click-through rate.

Specifically, we reuse the negative samples sampled for Equation \eqref{eq:l1}. For each quadruple $(u,v,r,w)\in \mathcal{O}$, we define the cross entropy loss function as follows.

\begin{equation}
\mathcal{L}_2 \! = \! -\frac{1}{\left|\mathcal{O}\right |}\left(
\sum_{\left (u,v,r,w \right )\in\mathcal{O}}\phi\left (r\right )\cdot\log{\sigma\left(g_2\left(\mathbf{z}_u,\mathbf{z}_{v}\right)\right)+\log{\left( 1-\sigma\left(g_2\left(\mathbf{z}_u,\mathbf{z}_{w}\right)\right)\right)}} \right)
\label{eq:l2}
\end{equation}

In Equation \eqref{eq:l2}, $g_2\left(\mathbf{z}_u,\mathbf{z}_{v}\right)=\mathbf{z}_u^T \mathbf{Q}_2 \mathbf{z}_{v}$ where $\mathbf{Q}_2$ is a learnable parameter, and $\phi(r)$ is used as the weight of the positive sample, and the weight of the negative sample is set to 1.

\noindent\textbf{Next click video prediction function}
User-video interactions can be viewed as a sequence which implies the change of user preferences on videos over time. Therefore, we can set the prediction of the next click video by the interaction sequences as an optimization objective. 

Given that a user has the interaction sequence $(v_1,v_2,\dots,v_T)\in \mathcal{S}$,  where $\mathcal{S}$ is the set of interaction sequences, for predicting the next click video, we apply the standard single layer GRU, taking $\mathbf{z}_{v_1},\mathbf{z}_{v_2},\dots,\mathbf{z}_{v_{T-1}}$ as input, and then regard the hidden state representation output by the GRU at the last time step as the embedding $\mathbf{q}_{v_T}$ of the video sequence. Next, we send  $\mathbf{q}_{v_T}$ to a fully connected layer and then apply softmax to predict the distribution of the next click video $v_{T}$. The corresponding equations are as follows.

\begin{equation}
\mathbf{q}_{v_T}={\rm GRU}({\mathbf z}_{v_1},{\mathbf z}_{v_2},\dots,{\mathbf z}_{v_{T-1})}
\label{eq:qt}
\end{equation}
\begin{equation}
p\left(v_T|v_1,v_2,\dots,v_{T-1} \right)={ softmax(\mathbf{W}_s\mathbf{q}_{v_T}})
\label{eq:p}
\end{equation}

In Equation \eqref{eq:p}, $\mathbf{W}_s^{|\mathcal{V}| \times d}$maps the sequence representation $\mathbf{q}_{v_T}$ to the $|\mathcal{V}|$ dimension space of the video set.
Furthermore, we take the negative log likelihood of $v_T$ as the function to be optimized, i.e., the sequence loss, as shown below.

\begin{equation}
\mathcal{L}_3=-\frac{1}{\left|\mathcal{S}\right |} \left(
\sum_{\left (v_1,v_2,\dots,v_T\right)\in \mathcal{S}}\log{p\left( v_T|v_1,v_2,\dots,v_{T-1} \right)} \right)
\label{eq:l3}
\end{equation}

\noindent\textbf{Loss function}
The loss function of the Sagittarius model is the weighted summation of the three optimization functions, as shown below.

\begin{equation}
\mathcal{L}=\lambda_1 \mathcal{L}_1+\lambda_2 \mathcal{L}_2+\lambda_3 \mathcal{L}_3
\label{eq:l}
\end{equation}

In Equation \eqref{eq:l}, $\lambda_1$,$\lambda_2$ and $\lambda_3$ denote the coefficients, which specify the importance of different functions, respectively. 

\subsection{Recommendation Acceleration}
% For the final top-k video recommendation, we apply the decoder $g_1(\mathbf{z}_u,\mathbf{z}_i)$ to predict the affinity score of each user-video pair $(u,i)$ and then sort by the affinity score to get top-k recommendation. The time complexity of generating recommendation results is $\mathcal{O}(|\mathcal{U}|\cdot|\mathcal{I}|)$. When the number of users is very large, just like the case in MX Player, it will take a lot of time to generate recommendation results if only using a single machine. 

% In order to speed up the generation of recommendation results, we execute the generation on Spark engine. The implementation steps of core algorithm are as follows:

When facing large-scale users, it will take a lot of time to generate recommendation results if only using a single machine. In order to speed up the generation of recommendation results, we perform the generation on Spark engine. The detailed implementation is as Algorithm \ref{alg:rec_algo}.
% \begin{algorithm}
% 	\caption{My algorithm}\label{euclid}  
% 	\begin{algorithmic}[1]  
% 		\Procedure{MyProcedure}{}  
% 		\State $\textit{stringlen} \gets \text{length of }\textit{string}$  
% 		\State $i \gets \textit{patlen}$ 
% 		\If {$i > \textit{stringlen}$} \Return false  
% 		\EndIf  
% 		\State $j \gets \textit{patlen}$
% 		\If {$\textit{string}(i) = \textit{path}(j)$}  
% 		\State $j \gets j-1$.  
% 		\State $i \gets i-1$.  
% 		\State \textbf{goto} \emph{loop}.  
% 		\State \textbf{close};  
% 		\EndIf  
% 		\State $i \gets i+\max(\textit{delta}_1(\textit{string}(i)),\textit{delta}_2(j))$.  
% 		\State \textbf{goto} \emph{top}.  
% 		\EndProcedure  
% 	\end{algorithmic}  
% \end{algorithm} 

\begin{algorithm}[h]
\caption{Generating top-k video recommendations}
\label{alg:rec_algo} 
    \begin{algorithmic}[1]
    \STATE 	According to the obtained embeddings $\mathbf{z}_u,\mathbf{z}_v$, construct the list $\mathbf{Y}$ in the form of $\left[ \left( u_1,\mathbf{z}_{u_1}\right),\dots, \left(u_{|\mathcal{U}|},\mathbf{z}_{u_{|\mathcal{U}|}}\right) \right]$, and form final video embedding matrix $\mathbf{Z} \in \mathbb{R}^{|\mathcal{V}| \times d}$.
    
    \STATE For each user $u\in\mathcal{U}$, obtain the historical interacted video collection $\mathcal{S}_u$, and then organize them into the form of $\left[ \left( u_1,\mathcal{S}_{u_1}\right),\dots, \left(u_{|\mathcal{U}|},\mathcal{S}_{u_{|\mathcal{U}|}}\right) \right]$.\
    
    \STATE Perform a concatenation operation of $\mathbf{Y}$ in Step (1) with the result of Step (2), and the result is like $\left[ (u,\mathbf{z}_{u},\mathcal{S}_{u}),\dots \right]$.\
    \STATE	Broadcast parameters $\mathbf{Z}$ and $\mathbf{Q}_1$ in 
    % the decoder 
    $g_1(\mathbf{z}_u,\mathbf{z}_v)
    =\mathbf{z}_u^T \mathbf{Q}_1 \mathbf{z}_v
    $ to each executor.\
    
    \STATE For each user $u \in \mathcal{U}$, this is, each $\left[ (u,\mathbf{z}_{u},\mathcal{S}_{u}),\dots \right]$ in the result of Step (3), perform the following Map operation:
    \begin{enumerate}
        \item[5.1] According to $\mathbf{z}_u$,$\mathbf{Z}$ and $\mathbf{Q}_1$, calculate the affinity score $g_1(\mathbf{z}_u,\mathbf{z}_v)$ of $u$ for each video $v \in \mathcal{V}$.
        % Traverse the video list which are sorted in descending order of the affinity score. For each video $i$, if $i \notin \mathcal{S}_u$ , it is added to the top-k recommendation result. When the number of recommended videos reaches $k$, the loop ends.
         \item[5.2] Traverse the set of videos which are sorted in descending order of the affinity score. For each video $v$, if $v \in \mathcal{S}_u$ then it is filtered, else it is added to the top-k recommendation result. When the number of recommended videos reaches $k$, the loop ends.
    \end{enumerate}
    
    \STATE Perform the Reduce operation and output the recommendation result of each user to the storage file (e.g., S3 in MX Player).
    \end{algorithmic}
\end{algorithm}

\section{Evaluation}

In this section, we evaluate the top-k video recommendation performance of Sagittarius by conducting offline experiments and online A/B tests. We list  corresponding results and discussions,  answering the following four research questions (RQs):

\textbf{RQ1:} How well does Sagittarius perform for the top-k video recommendation, compared to the state-of-the-art GNN models? 

\textbf{RQ2:} What is the impact of the design choices of Sagittarius on the performance of the top-k video recommendation?

\textbf{RQ3:} What is the impact of the hyper-parameters of Sagittarius on the recommendation performance?

\textbf{RQ4:} How does Sagittarius perform in the live production environment, e.g., when serving MX Player, one of India's largest streaming platforms?

\subsection{Experimental Setup}
\textbf{Datasets} We adopt four publicly available datasets, i.e., MovieLens-100K and MovieLens-10M from MovieLens \cite{movielens}, and Amazon-Beauty and Amazon-Digital Music from Amazon Product Review\cite{amazon}. For the latter two datasets, we filter out items which are reviewed less than five times and users who give less than five reviews. For these four datasets, we split them into train, validation and test sets by a ratio of nearly 7:1:2.

Besides, we construct a dataset MXPlayer-4D-5M from the MX Player log, which contains four-day data from Oct. 31st, 2020 to Nov. 3rd, 2020. We take the data in the first three days as a train set from which we take 10\% as a validation set, and then take the data of the last day as a test set.

Table 1 lists the statistics of these five datasets. Roughly, we can classify datasets by density into dense and sparse datasets, where MovieLens-100K and MovieLens-10M belong to the former and Amazon-Beauty, Amazon-Digital Music and MXPlayer-4D-5M datasets belong to the latter. In addition, as shown in the last column of Table 1, we set different $\phi(r)$ for different datasets. Taking MXPlayer-4D-5M as an example, we set $\phi(r)$ to 0.5, 1.0, 1.5, 2.0, 2.5, 3.0, 3.5, 4.0, 4.5 and 5 for user’s single click, multiple clicks, watching whose duration is between 10s and 1min, watching whose duration is between 1min and 5min, favorite, watching whose duration is between 5min and 30min, watching whose duration is greater than 30min, sharing, like, and download, respectively.

\begin{table}[ht]
	\centering
	\label{tab:stat}
	\caption{Statistics of Datasets}
	\setlength{\tabcolsep}{2mm}
	\scriptsize 
	\begin{tabular}{lrrrrr}
		\toprule
		Dataset & \#User & \#Item & \#Ratings & Density & Rating Levels \\
		\midrule
		MovieLens-100K & 943 & 1682 & 100,000 & 0.0630 & $1, 2, \ldots, 5$ \\
		MovieLens-10M & 69,878 & 10,677 & 10,000,054 & 0.0134 &$0.5, 1, 1.5, 2 \ldots, 5$ \\
		Amazon-Beauty  & 12,008 & 3,570 & 92,512 & 0.0022 & $1, 2, \ldots, 5$\\
		Amazon-Digital Music & 4,325 & 1,662 & 38,722 & 0.0054 &$1, 2, \ldots, 5$  \\
		MXPlayer-4D-5M & 5,534,825 & 13,463 & 29,471,665 & 0.0004 &$0.5, 1, 1.5, 2 \ldots, 5$ \\
		\bottomrule
	\end{tabular}
\end{table}

\noindent\textbf{Metrics} We adopt Recall@K, URecall@K (Unique Recall@K) and NDCG@K (Normalized Discounted Cumulative Gain@K) as evaluation metrics. As for URecall@K, we say that for a user, if there is at least one positive sample among the top-k recommended items, then the URecall@K is 1, otherwise it is 0. The Unique Recall@K of the recommendation system is the average of Unique Recall@K values of each user. That is, Unique Recall@K is equivalence to Hit@K. 

\subsection{Competitors }
For each dataset, we search the optimal parameters for each model using the train set and validation set, and then conduct the comparative experiments on the test set using the models under optimal parameters.

We choose the following state-of-the-art models which are built on bipartite user-item interaction graphs as the competitors: 

\begin{enumerate}
    \item \textbf{GCMC}\cite{berg2018graph}: a graph auto-encoder which predicts labeled links in the  bipartite user-item graph. To apply GCMC to our scenario, we treat different user behaviors in our scenario as different labeled links in GCMC.
    
    \item \textbf{NGCF}\cite{wang2019neural}: a GCN based method which refines embeddings of users and items by embedding propagation.
    
    \item \textbf{LightGCN}\cite{he2020lightgcn}: a simplified version of NGCF which omits the feature transformation and nonlinear activation. 
    
    \item \textbf{MMGCN}\cite{wei2019mmgcn}: a GCN based method which generates separate node embeddings for each modality and then combines them into the final ones. To apply MMGCN to our scenario, we treat each category of user behaviors in our scenario as a modality individually. 

\end{enumerate}

To be fair, for Sagittarius and all the above methods, we adopt 2-layer convolutional neural networks, set the dimensions of user and video embedding to 64, and set the number of negative samplings in BPR loss function to 10. We adopt Adam optimizer, setting the learning rate to 0.01.

\subsection{Performance Comparison}
We conduct performance comparison experiments, comparing Sagittarius with four competitors. Table \ref{tab:performance} lists the recommendation performance of Sagittarius and all the competitors.

\begin{table}[ht]
	\centering
	\caption{Recommendation Performance. The best performance in each row is in bold, and the second best performance in each row is underlined.}
	\scriptsize
	\label{tab:performance}
	\begin{tabular}{llcccccc}
		\toprule
		Datasets  &  Metrics & GCMC  & NGCF    & LightGCN  & MMGCN    & Sagittarius &  Improvement \\
		\midrule
		\multirow{2}{*}{MovieLens-100K}  
		& \textbf{Recall@10} & 0.0356	& 0.0361	& \underline{0.0367}	& 0.0343	& \textbf{0.0382}	& 4.09\%\\
		& \textbf{URecall@10} & 0.4542	& 0.4657	& \underline{0.4683}	& 0.4497	& \textbf{0.4751}	& 1.45\%\\
		& \textbf{NDCG@10} & 0.3207	& 0.3219	& \underline{0.3285}	& 0.3108	& \textbf{0.3392}	& 3.26\%\\
		\multirow{2}{*}{MovieLens-10M}  
		& \textbf{Recall@10}  & 0.0068	& \underline{0.0074}	& 0.0073	& 0.0072	& \textbf{0.0076}    & 2.70\%\\
		& \textbf{URecall@10} & 0.4651	& \underline{0.5081}	& 0.5058	& 0.4981	& \textbf{0.5142}	& 1.20\%\\
		& \textbf{NDCG@10} & 0.2502	& \underline{0.2863}	& 0.2761	& 0.2714	& \textbf{0.2963}	& 3.49\%\\
		\multirow{2}{*}{Amazon-Beauty} 
		& \textbf{Recall@10}  & 0.0362	& 0.0340	& 0.0357	& \underline{0.0372}	& \textbf{0.0445}	& 19.62\%\\
		& \textbf{URecall@10} & 0.0661	& 0.0652	& 0.0658	& \underline{0.0681}	& \textbf{0.0768}	& 12.78\%\\
		& \textbf{NDCG@10} & 0.0573	& 0.0562	& 0.0571	& \underline{0.0587}	& \textbf{0.0604}	& 2.90\%\\
		\multirow{2}{*}{Amazon-Digital Music}
		& \textbf{Recall@10} & 0.0257	& 0.0251	& 0.0254	& \underline{0.0265}	& \textbf{0.0311}	& 17.36\%\\
		& \textbf{URecall@10} & \underline{0.0723}	& 0.0701	& 0.0716	& 0.0713	& \textbf{0.0815}	& 12.72\%\\
		& \textbf{NDCG@10} & 0.0995	& 0.0887	& 0.0913	& \underline{0.1004}	& \textbf{0.1016}	& 1.20\%\\
		\multirow{2}{*}{MXPlayer-4D-5M}
		& \textbf{Recall@10} & 0.1653	& 0.1539	& 0.1367	& \underline{0.1710}	& \textbf{0.1794}	& 4.91\%\\
		& \textbf{URecall@10} & 0.2975	& 0.2913	& 0.2748	& \underline{0.3014}	& \textbf{0.3147}	& 4.41\%\\
		& \textbf{NDCG@10} & 0.2139	& 0.2094	& 0.1985	& \underline{0.2172}	& \textbf{0.2261}	& 4.10\%\\
		\bottomrule
	\end{tabular}
\end{table}

From Table \ref{tab:performance}, we can make the following observations and inferences:
\begin{enumerate}
    \item On dense datasets, Sagittarius, LightGCN and NGCF outperform GCMC and MMGCN. However, on sparse datasets, Sagittarius and MMGCN behave better than GCMC, LightGCN and NGCF, which shows that, in the scenarios with sparse datasets, differentiating between different interactive behaviors and model them separately are effective.
    \item LightGCN does not perform as well as NGCF on datasets except MovieLens-100K, Amazon-Beauty and Amazon-Digital Music. These results are different from the ones in \cite{he2020lightgcn}, which illustrate that deleting the feature transformation and nonlinear activation will not always obtain the stable improvement in performance and the actual effect of LightGCN might depend on the datasets or scenarios.
    \item Whether it is for dense datasets or sparse datasets, Sagittarius outperforms other comparison models in all metrics. The reasons can be summarized as follows. Firstly, Sagittarius quantifies interactions of users on items and applies the quantitative values to the messages needed to be propagated in the GCN. Secondly, Sagittarius quantifies interaction behaviors in weights to guide the objective function to pay more attention to samples with the high interaction value. Thirdly, Sagittarius adopts a combination of multiple optimization objectives to tradeoff the relationship between different objectives, thereby fully mining the information in the interactions and improving the concentration level of the model.
\end{enumerate}

\subsection{Ablation Analyses}

We conduct the ablation study to observe the effectiveness of different components in Sagittarius including three optimization objectives (denoted as CTR, Sequence, BPR, respectively), behavior weighting (denoted as Behavior). Table \ref{tab:ablation} shows the detailed results of the ablation study.

\begin{table}[ht]
	\centering
	\caption{Ablation study of Sagittarius. The best performance in each row is the number in bold, and the worst performance in each row is underlined.}
	\scriptsize
	\label{tab:ablation}
	\setlength{\tabcolsep}{2mm}
	\begin{tabular}{clccccc}
		\toprule
		Datasets  &  Metrics & Sagittarius  & - CTR    & - Sequence  & - BPR    & - Behavior\\
		\midrule
		\multirow{2}{*}{MovieLens-100K}  
		& \textbf{Recall@10} & \textbf{0.0382}	& 0.0376	& 0.0374	& \underline{0.0362} 	& 0.0377\\
		& \textbf{URecall@10} & \textbf{0.4751}	& 0.4687	& 0.4652	& \underline{0.4575} 	& 0.4691\\
		& \textbf{NDCG@10} & \textbf{0.3392}	& 0.3314	& 0.3279	& \underline{0.3217} 	& 0.3354\\
		\multirow{2}{*}{MovieLens-10M}  
		& \textbf{Recall@10}  & \textbf{0.0076}	& 0.0074 	& 0.0072 	& \underline{0.0069} 	& 0.0074\\
		& \textbf{URecall@10} & \textbf{0.5142}	& 0.5068	& 0.4976	& \underline{0.4753} 	& 0.5117\\
		& \textbf{NDCG@10} & \textbf{0.2963}	& 0.2907	& 0.2883	& \underline{0.2648} 	& 0.2935\\
		\multirow{2}{*}{Amazon-Beauty} 
		& \textbf{Recall@10}  & \textbf{0.0445}	& 0.0423	& 0.0378	& \underline{0.0365} 	& 0.0382\\
		& \textbf{URecall@10} & \textbf{0.0768}	& 0.0751	& 0.0739	& \underline{0.0724} 	& 0.0735\\
		& \textbf{NDCG@10} & \textbf{0.0604}	& 0.0591	& 0.0579	& \underline{0.0571} 	& 0.0587\\
		\multirow{2}{*}{\shortstack{Amazon-Digital \\ Music}}
		& \textbf{Recall@10}  & \textbf{0.0311}	& 0.0295	& 0.0274 	& \underline{0.0254}    & 0.0287\\
		& \textbf{URecall@10} & \textbf{0.0815}	& 0.0783	& 0.0746	& \underline{0.0703} 	& 0.0765\\
		& \textbf{NDCG@10} & \textbf{0.1016}	& 0.0967	& 0.0912	& \underline{0.0897} 	& 0.0924\\
		\multirow{2}{*}{MXPlayer-4D-5M}
		& \textbf{Recall@10} & \textbf{0.1794}	& 0.1643	& 0.1624	& \underline{0.1452} 	& 0.1712\\
		& \textbf{URecall@10} & \textbf{0.3147}	& 0.2916	& 0.2937	& \underline{0.2758} 	& 0.3024\\
		& \textbf{NDCG@10} & \textbf{0.2261}	& 0.2105	& 0.2088	& \underline{0.1936} 	& 0.2193\\
		\bottomrule
	\end{tabular}
\end{table}

From Table \ref{tab:ablation}, we find that removing the BPR optimization objective leads to the largest drop in performance metrics and no matter which optimization objective is removed, the performance decreases on all datasets, which shows that combining multiple optimization objectives actually can help improve performance.

On the other hand, on the sparse datasets, the performance of Sagittarius-behavior which removes the weighting of interaction behavior is not as good as the one of intact Sagittarius, which is our expectation. On the dense dataset, the performance of the Sagittarius-behavior has not much difference from the intact model. This shows that on sparse datasets, the behavior-aware strategy, i.e., differentiating from interaction behaviors and converting interaction behaviors into weights to guide learning objectives to pay more attention to high-affinity user-item pairs, can indeed improve recommendation performance.

\subsection{Impact of Hyper-parameters}
In this section, we analyze the impact of hyper-parameters. Due to the limit of space, we only select three important hyper-parameters, that is, $\lambda_1$, $\lambda_2$, and $\lambda_3$, and conduct experiments to observe the performance under different values of $\lambda_1$, $\lambda_2$, and $\lambda_3$.

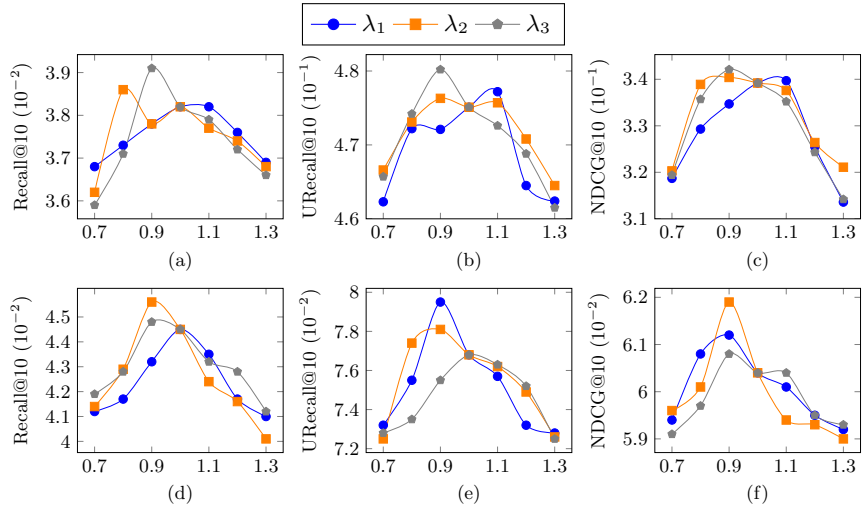
\begin{figure}[H]
\centering
\pgfplotsset{footnotesize}
\begin{center}
\ref{named}\\
%note that \centering uses less vspace...
\begin{tikzpicture}[xscale=0.8, yscale=0.8]

\begin{axis}[
legend columns=-1,
legend entries={$\lambda_1$,$\lambda_2$,$\lambda_3$,},
legend to name=named, 
xlabel=(a),
ylabel=Recall@10 ($10^{-2}$),
xtick={0.7,0.9,1.1,1.3},
ytick={3.6, 3.7, 3.8,3.9,4.0},
% ymin=0.0345, 
% ymax=0.040,    
% scaled ticks=false, 
yticklabel style={
/pgf/number format/precision=3,   
},],
\addplot+[smooth,color=blue,mark=*]
        plot coordinates {
            (0.7,3.68)
            (0.8,3.73)
            (0.9,3.78)
            (1.0,3.82)
            (1.1,3.82)
            (1.2,3.76)
            (1.3,3.69)
        };
\addplot+[smooth,color=orange,mark=square*]
        plot coordinates {
            (0.7,3.62)
            (0.8,3.86)
            (0.9,3.78)
            (1.0,3.82)
            (1.1,3.77)
            (1.2,3.74)
            (1.3,3.68)
        };
\addplot+[smooth,color=gray,mark=pentagon*]
        plot coordinates {
            (0.7,3.59)
            (0.8,3.71)
            (0.9,3.91)
            (1.0,3.82)
            (1.1,3.79)
            (1.2,3.72)
            (1.3,3.66)
        };
\end{axis}
\end{tikzpicture}
\begin{tikzpicture}[xscale=0.8, yscale=0.8]
\begin{axis}[
legend columns=-1,
legend entries={$\lambda_1$,$\lambda_2$,$\lambda_3$,},
legend to name=named,
xlabel=(b),
ylabel=URecall@10 ($10^{-1}$),
xtick={0.7,0.9,1.1,1.3},
ytick={4.60,4.7,4.80,4.9},
ymin=4.6,
% scaled ticks=false, %取消科学计数法
yticklabel style={
/pgf/number format/precision=3,    % 三位有效数字
},
]
\addplot+[smooth,color=blue,mark=*]
        plot coordinates {
            (0.7,4.623)
            (0.8,4.722)
            (0.9,4.721)
            (1.0,4.751)
            (1.1,4.772)
            (1.2,4.645)
            (1.3,4.624)
        };
\addplot+[smooth,color=orange,mark=square*]
        plot coordinates {
            (0.7,4.666)
            (0.8,4.731)
            (0.9,4.763)
            (1.0,4.751)
            (1.1,4.757)
            (1.2,4.708)
            (1.3,4.645)
        };
\addplot+[smooth,color=gray,mark=pentagon*]
        plot coordinates {
            (0.7,4.657)
            (0.8,4.742)
            (0.9,4.802)
            (1.0,4.751)
            (1.1,4.726)
            (1.2,4.688)
            (1.3,4.615)
        };
\end{axis} 
\end{tikzpicture} 
\begin{tikzpicture}[xscale=0.8, yscale=0.8]
\begin{axis}[
legend columns=-1,
legend entries={$\lambda_1$,$\lambda_2$,$\lambda_3$,},
legend to name=named,
xlabel=(c),
ylabel=NDCG@10 ($10^{-1}$),
xtick={0.7,0.9,1.1,1.3},
ytick={3.1,3.2,3.3,3.40},
ymin=3.1,
% scaled ticks=false,
yticklabel style={
/pgf/number format/precision=3,    
},]

\addplot+[smooth,color=blue,mark=*]
        plot coordinates {
            (0.7,3.187)
            (0.8,3.293)
            (0.9,3.347)
            (1.0,3.392)
            (1.1,3.397)
            (1.2,3.255)
            (1.3,3.136)
        };
\addplot+[smooth,color=orange,mark=square*]
        plot coordinates {
            (0.7,3.203)
            (0.8,3.389)
            (0.9,3.404)
            (1.0,3.392)
            (1.1,3.376)
            (1.2,3.264)
            (1.3,3.211)
        };
\addplot+[smooth,color=gray,mark=pentagon*]
        plot coordinates {
            (0.7,3.194)
            (0.8,3.357)
            (0.9,3.421)
            (1.0,3.392)
            (1.1,3.352)
            (1.2,3.243)
            (1.3,3.142)
        };
\end{axis} 
\end{tikzpicture}\\

\begin{tikzpicture}[xscale=0.8, yscale=0.8]
\begin{axis}[
xlabel=(d),
ylabel=Recall@10 ($10^{-2}$),
xtick={0.7,0.9,1.1,1.3},
ytick={4.0,4.1,4.2,4.3,4.4,4.5},
% scaled ticks=false, 
yticklabel style={
/pgf/number format/precision=3,  
},]

\addplot+[smooth,color=blue,mark=*]
        plot coordinates {
            (0.7,4.12)
            (0.8,4.17)
            (0.9,4.32)
            (1.0,4.45)
            (1.1,4.35)
            (1.2,4.17)
            (1.3,4.1)
        };
\addplot+[smooth,color=orange,mark=square*]
        plot coordinates {
            (0.7,4.14)
            (0.8,4.29)
            (0.9,4.56)
            (1.0,4.45)
            (1.1,4.24)
            (1.2,4.16)
            (1.3,4.01)
        };
\addplot+[smooth,color=gray,mark=pentagon*]
        plot coordinates {
            (0.7,4.19)
            (0.8,4.28)
            (0.9,4.48)
            (1.0,4.45)
            (1.1,4.32)
            (1.2,4.28)
            (1.3,4.12)
        };
\end{axis} 
\end{tikzpicture}
\begin{tikzpicture}[xscale=0.8, yscale=0.8]
\begin{axis}[
legend columns=-1,
legend entries={$\lambda_1$,$\lambda_2$,$\lambda_3$,},
legend to name=named,
xlabel=(e),
ylabel=URecall@10 ($10^{-2}$),
xtick={0.7,0.9,1.1,1.3},
ytick={7.2,7.4,7.6,7.8,8.0},
% scaled ticks=false,
yticklabel style={
/pgf/number format/precision=3,   
},]
						
\addplot+[smooth,color=blue,mark=*]
        plot coordinates {
            (0.7,7.32)
            (0.8,7.55)
            (0.9,7.95)
            (1.0,7.68)
            (1.1,7.57)
            (1.2,7.32)
            (1.3,7.28)
        };
\addplot+[smooth,color=orange,mark=square*]
        plot coordinates {
            (0.7,7.25)
            (0.8,7.74)
            (0.9,7.81)
            (1.0,7.68)
            (1.1,7.62)
            (1.2,7.49)
            (1.3,7.26)
        };
\addplot+[smooth,color=gray,mark=pentagon*]
        plot coordinates {
            (0.7,7.28)
            (0.8,7.35)
            (0.9,7.55)
            (1.0,7.68)
            (1.1,7.63)
            (1.2,7.52)
            (1.3,7.25)
        };
\end{axis} 
\end{tikzpicture}
\begin{tikzpicture}[xscale=0.8, yscale=0.8]
\begin{axis}[
legend columns=-1,
legend entries={$\lambda_1$,$\lambda_2$,$\lambda_3$,},
legend to name=named,
xlabel=(f),
ylabel=NDCG@10 ($10^{-2}$),
xtick={0.7,0.9,1.1,1.3},
ytick={5.9,6.0,6.1,6.2},
% scaled ticks=false, 
yticklabel style={
/pgf/number format/precision=3,  
},]
						
\addplot+[smooth,color=blue,mark=*]
        plot coordinates {
            (0.7,5.94)
            (0.8,6.08)
            (0.9,6.12)
            (1.0,6.04)
            (1.1,6.01)
            (1.2,5.95)
            (1.3,5.92)
        };
\addplot+[smooth,color=orange,mark=square*]
        plot coordinates {
            (0.7,5.96)
            (0.8,6.01)
            (0.9,6.19)
            (1.0,6.04)
            (1.1,5.94)
            (1.2,5.93)
            (1.3,5.9)
        };
\addplot+[smooth,color=gray,mark=pentagon*]
        plot coordinates {
            (0.7,5.91)
            (0.8,5.97)
            (0.9,6.08)
            (1.0,6.04)
            (1.1,6.04)
            (1.2,5.95)
            (1.3,5.93)
        };
\end{axis} 
\end{tikzpicture}
\end{center}
\caption{The influence of  $\lambda_1$, $\lambda_2$ and $\lambda_3$ on MovieLens-100K and Amazon-Beauty} \label{fig:lambda}
\end{figure}

For the convenience of experiments, we change one of three parameters from 0.7 to 1.3 with the increment of 0.1 while setting the rest parameters to 1. Figures \ref{fig:lambda}(a)-(c) show the results on MovieLens-100K and Figures \ref{fig:lambda}(d)-(f) show the results on Amazon-Beauty, where all the blue lines with solid circles show the performance under different $\lambda_1$ but $\lambda_2=1$ and $\lambda_3=1$, all the orange lines with solid squares show the performance under different $\lambda_2$ but $\lambda_1=1$ and $\lambda_3=1$, and all the grey lines with solid pentagons show the performance under different $\lambda_3$ but $\lambda_1=1$ and $\lambda_2=1$. 

From Figure \ref{fig:lambda}, we find no matter which parameter is adjusted, the change trends of different performance metrics are the same, that is, with the change of any hyper-parameter value from small to large, the performance varies from a low value to a high one and finally moves to a low one. This observation illustrates that substantially increasing or decreasing the parameter value to emphasize or weaken some objective will slow down the performance. In addition, as shown in Figure \ref{fig:lambda}, when values of three parameters are within  $[0.9, 1.1]$, the performance is relatively good. Therefore, we assign the same value (i.e., 1) to these three parameters. 

% Preamble: \pgfplotsset{width=7cm,compat=1.15}

\subsection{Online A/B Test}

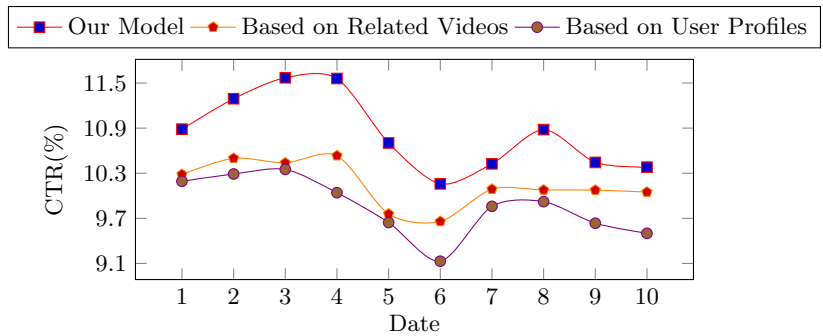
\begin{figure}[H]
\centering

\begin{tikzpicture}
    \pgfplotsset{ 
        every axis legend/.append style={ at={(0.5,1.05)}, anchor=south }, 
        every axis x label/.style={at={(0.5,0)},below,yshift=-10pt},
        every axis y label/.style={at={(0,0.5)},xshift=-30pt,rotate=90}
        } 
    \begin{axis}[
        % title=123,
        % legend cell align=center, 
        % legend pos=outer north east,
        % xmin=0, xmax=12,
        y=1cm,
        legend columns=3,
        xlabel=Date,
        ylabel=CTR(\%),
        xtick={1,2,3,4,5,6,7,8,9,10},
        ytick={9.1,9.7,...,11.5}],
        
        \addplot+[smooth,color=red,mark=square*]
        plot coordinates {

            (1,10.885831)
            (2,11.2912863)
            (3,11.5703657)
            (4,11.5601927)
            (5,10.702461)
            (6,10.1584569)
            (7,10.4250476)
            (8,10.8806774)
            (9,10.4438528)
            (10,10.3803657)

        };
    \addlegendentry{Case 1}
    \addplot+[smooth,color=orange,mark=pentagon*]
        plot coordinates {
            (1,10.2849461)
            (2,10.4990296)
            (3,10.4379326)
            (4,10.5334893)
            (5,9.7583733)
            (6,9.6584342)
            (7,10.0883916)
            (8,10.0775145)
            (9,10.0754142)
            (10,10.0490384)
        };
    \addlegendentry{Case 2}
    \addplot+[smooth,color=violet,mark=*]
        plot coordinates {
            (1,10.1926379)
            (2,10.2898434)
            (3,10.3508651)
            (4,10.0405119)
            (5,9.6441351)
            (6,9.1293938)
            (7,9.8596662)
            (8,9.9215269)
            (9,9.63507)
            (10,9.5009178)
        };
    \addlegendentry{Case 2}

    \legend{Our Model,Based on Related Videos,Based on User Profiles}
    \end{axis}
\end{tikzpicture}
\caption{Online A/B Test (Sep. 13th, 2020 $\sim$  Sep. 23th, 2020)} \label{fig:abtest}
\end{figure}

We have deployed Sagittarius in the production environment of MX Player to serve top-k video recommendation scenarios. We observe the difference in performance between Sagittarius and two existing recommendation models in MX Player. One comparison model is to recommend top-k videos based on the videos clicked by users recently, and the other one is to recommend videos according to user profiles. Figure \ref{fig:abtest} shows the CTRs of three models recommending movie-type videos from Sep. 13rd, 2020 to Sep. 23rd, 2020.

From Figure \ref{fig:abtest}, we can see that the CTRs of Sagittarius are significantly better than the existing recommendation models. During the ten-day observation, Sagittarius is always in the first place, which shows the effectiveness of Sagittarius. Currently, the recommendation results obtained by Sagittarius are used to generate user’s personalized card. This personalized card is entitled “Movies Based on Your Viewing” and shown on the start screen in the MX Player App.

\section{Conclusion}

To improve the performance of top-k video recommendation in MX Player, we propose a model named Sagittarius in the paper. Sagittarius extracts the collaborative relations in the  bipartite user-item graph to the node embeddings through the convolution layers. Meanwhile, in Sagittarius, we quantify the behaviors in the user-video interactions and apply them to guide the message propagation and the optimization of the recommendation task. More importantly, for the top-k video recommendation, we propose to choose the videos by multiple metrics and adopt a combination of three optimization objects to drive the training of the model. Results from offline experiments and online A/B tests illustrate that Sagittarius is suitable for top-k recommendation scenarios with sparse data, in addition to MX Player.





% LNCS Homepage, \url{http://www.springer.com/lncs}. Last accessed 4
% \end{thebibliography}
\bibliographystyle{splncs04}
\bibliography{samplepaper}
\end{document}